\documentclass[conference]{IEEEtran}
\IEEEoverridecommandlockouts
\usepackage{cite}
\usepackage{amsmath,amssymb,amsfonts}
\usepackage{graphicx}
\usepackage{textcomp}
\usepackage{xcolor}
\usepackage{bm}
\usepackage[ruled,vlined]{algorithm2e}
\usepackage{makecell}
\usepackage{graphicx}
\usepackage{textcomp}
\usepackage{array} 
\usepackage{longtable}
\usepackage{booktabs}
\usepackage{float}
\usepackage{diagbox}
\usepackage{subfig}
\usepackage{multirow}
\usepackage[hidelinks]{hyperref}
\usepackage{color}
\def\BibTeX{{\rm B\kern-.05em{\sc i\kern-.025em b}\kern-.08em
    T\kern-.1667em\lower.7ex\hbox{E}\kern-.125emX}}
\begin{document}

\title{Motion Robust High-Speed Light-Weighted Object Detection with Event Camera}

\author{Bingde Liu, Chang Xu, Wen Yang, Huai Yu, Lei Yu
\thanks{Manuscript received 8 January 2022; revised 5 March 2023; accepted 5 April 2023. Date of publication XXX 2023; date of current version XXX 2023. This work was supported in part by the National Natural Science Foundation of China under Grants 62271354; in part by the Natural Science Foundation of Hubei Province, China under Grants 2022CFB600 and 2021CFB467. The Associate Editor coordinating the review process was Dr. Emanuele Zappa. (Corresponding author: Wen Yang.)

The authors are with the School of Electronic Information, Wuhan University, Wuhan 430072, China. (e-mail: \{liubingde, xuchangeis, yangwen, yuhuai, ly.wd\}@whu.edu.cn)} 
}

\maketitle

\begin{abstract}
The event camera asynchronously produces the event stream with a high temporal resolution, discarding redundant visual information and bringing new possibilities for moving object detection. Nevertheless, the existing object detectors cannot make the most of the spatial-temporal asynchronous nature and high temporal resolution of the event stream. For one thing, existing methods fail to consider objects with different velocities relative to the event camera's motion, resulting from the global synchronized time window with the whole spatial slice. 
For another, most of the existing methods rely on heavy models and boost the detection performance with low frame rates, failing to utilize the high temporal resolution characteristic of the event stream. 
In this work, we propose a motion robust and high-speed detection pipeline which better leverages the event data. First, we design an event stream representation called Temporal Active Focus (TAF), which efficiently utilizes the spatial-temporal asynchronous event stream, constructing event tensors robust to object motions.
Then, we propose a module called the Bifurcated Folding Module (BFM), which encodes the rich temporal information in the TAF tensor at the input layer of the detector. 
Following this, we design a high-speed lightweight detector called Agile Event Detector (AED) plus a simple but effective data augmentation method, to enhance the detection accuracy and reduce the model's parameter. 
Experiments on two typical real-scene event camera object detection datasets show that our method is competitive in terms of accuracy, efficiency, and the number of parameters. By classifying objects into multiple motion levels based on the optical flow density metric, we further illustrated the robustness of our method for objects with different velocities relative to the camera. The codes and trained models are available at https://github.com/HarmoniaLeo/FRLW-EvD.
\end{abstract}


\section{Introduction}

Real-world object detection tasks have extremely high requirements on the detection speed and the robustness of detectors for bad weather, extreme lighting conditions, and fast-moving objects. 
Autonomous driving is one of its typical application scenarios which is demanding for safety and robustness.
Therefore, diverse sensors are leveraged to obtain robust detection under various conditions. Among them, traditional frame-based RGB cameras~\cite{yolov4_5d_2021_tim,survey_autonomous_tits_2020} are used to provide rich semantic and texture information about the surroundings. The LIght Detection And Ranging (LIDAR) sensor-based object detectors~\cite{pulsed3dlidars_2021_tim, efficient3dlidar_2022_tim, multibeamlidar_2021_tim, mv3d_2017_cvpr} are employed to compensate for RGB cameras' failure under extreme lighting conditions. In addition, the millimeter-wave radar is utilized to enhance the detection performance under adverse weather conditions~\cite{mrpt_2021_tim}. 
Despite these progresses, none of these sensors are particularly good at seeing fast-moving objects~\cite{pagad2020robust_icra,survey_autonomous_tits_2020} (\textit{e.g.} the sudden appearing pedestrian in front of a car), thus degrading the detection accuracy, which leads to a safety hazard.

As a new type of vision-based sensor~\cite{visioncamera_tim_2014}, the event camera has many advantages, including high temporal resolution, large dynamic range, and reduction of redundant information~\cite{serrano2013128,2008A,finateu20205,cao2022neurograsp}. Benefiting from these properties, event cameras have great potential for applications in scenarios where most sensors (\textit{e.g.} RGB camera, LIDAR, and millimeter-wave radar) are subject to motion blur and fast-moving objects, with very low energy consumption and small data storage costs~\cite{posch2014retinomorphic,gallego2020event}.

So far, there have been some works that apply event cameras to object detection tasks, mainly in the field of autonomous driving \cite{chen2020event, maqueda2018event, binas2017ddd17}. The contributions of the event cameras to the field are their ability to supplement other sensors, provide concurrent streams of temporal contrast events, and offer several advantages over traditional cameras. 

\begin{figure}[tbp]
\centering
\includegraphics[width=\linewidth]{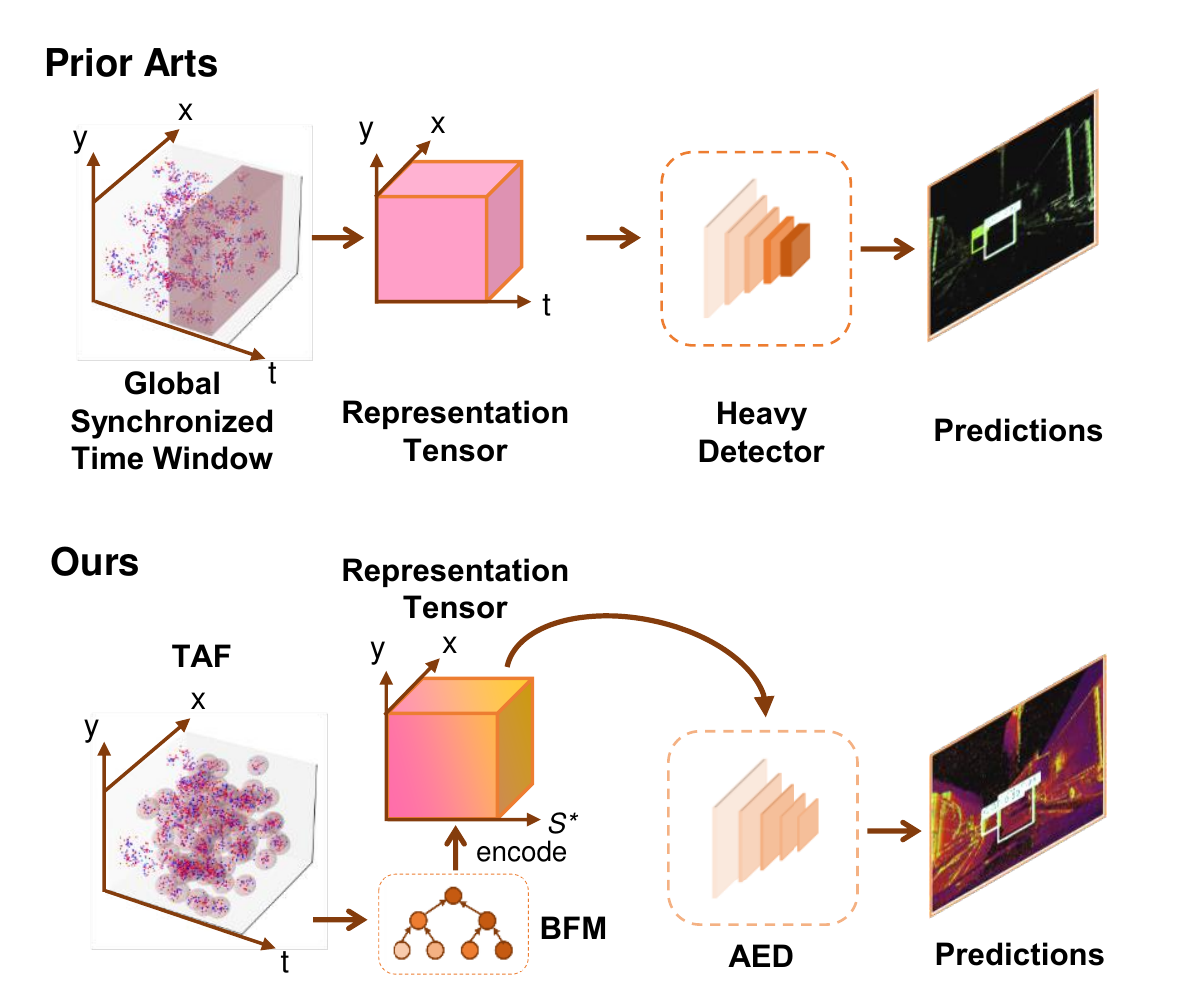}
\caption{Structure of our approach and comparison with previous approaches\cite{chen2018pseudo, li2019event,  perot2020learning, li2022asynchronous}. The TAF, BFM, and AED stand for the temporal active focus representation, bifurcated folding module, and agile event detector, respectively. \textit{S}* denotes the semantic dimension encoded via the BFM from the temporal dimension. Unlike previous methods, our TAF approach uses an asynchronous approach to sample the event stream and uses BFM to extract semantics from the temporal dimension when constructing the representation tensor. Our AED detector is also more lightweight.}
\label{fig:structure}
\end{figure}

Among the works, the methods with leading accuracy levels (\textit{e.g.,}~YOLE~\cite{canniciasynchronous}, RED~\cite{perot2020learning}, and ASTMNet~\cite{li2022asynchronous}) follow the paradigm of a global time window (cuboid-shaped shadow region in Fig.~\ref{fig:structure}) for event representation and a deep neural network-based object detector with large amounts of parameters to boost detection accuracy, their general structure is shown in Fig.~\ref{fig:structure}. For example, the state-of-the-art ASTMNet~\cite{li2022asynchronous} deeply explores this paradigm by adaptively adjusting the duration of the global time window and leveraging the recurrent-convolutional architecture.

Despite the exploration of adaptive and asynchronous temporal sampling, there still exist two main drawbacks in existing methods, namely the global time window and the unsatisfactory running speed~\cite{chen2018pseudo,li2022asynchronous,perot2020learning,canniciasynchronous,messikommer2020event,li2019event}. 
Firstly, the asynchronous exploration in existing methods is limited by the global time window, which maintains the whole spatial slice when adjusting the temporal duration. 
Hence, existing methods cannot take into account objects with different velocities relative to the event camera. When there are multiple objects in the field of view at the same time, a long time window may lead to the occurrence of motion blur for fast-moving objects, while a short time window may be unable to gather enough information for slow-moving objects. 
Secondly, current object detectors require a large number of model parameters to achieve high accuracy, which implies a large computational burden and low inference speed. Applying those algorithms to event data does not match its high temporal resolution nature. 

To solve the above problems, we propose a motion-robust and high-speed object detection pipeline for event data. First of all, we design a new event stream representation called Temporal Active Focus (TAF), to conquer the drawbacks of the global time window. It can better take advantage of the asynchronous nature of event data, spatially and temporally. By asynchronously sampling the Event Measurement Field \cite{gehrig2019end} in a low computationally demanding manner, it constructs tensors containing different time range information on different spatial and polar positions. Then, we design a module called Bifurcated Folding Module (BFM), cooperating with the TAF. The BFM encodes the rich temporal information in TAF tensors before feeding the tensors into the detector, improving the detection accuracy. Moreover, we propose a lightweight detector called Agile Event Detector (AED), which has a high inference speed that better matches the event data's high temporal resolution. In addition, to improve the generalization capability of the detector, we propose to use a simple but effective data augmentation strategy including random flipping and cropping. The overall architecture of our approach and comparison with the previous approach is shown in Fig. \ref{fig:structure}. 

We choose two typical real-scene event camera object detection datasets for experiments: the complete Prophesee GEN1 Automotive Detection Dataset (GEN1 Dataset)\cite{de2020large} and the Prophesee 1 MEGAPIXEL Automotive Detection Dataset \cite{perot2020learning} with partial annotation (1 MEGAPIXEL Dataset (Subset)). We measure the motion speed of objects in the datasets by computing optical flow, and then we classify them into 5 different motion levels. Experiments show that compared with the current state-of-the-art methods, our method has a far lower number of parameters and much higher running speed while retaining competitive accuracy. The experiments also show that our method has high detection accuracy under all 5 motion levels, which demonstrates motion robustness. 

In summary, our contributions are as follows: 

\begin{enumerate}
    \item We propose an event stream representation method called Temporal Active Focus (TAF), which better leverages the asynchronous nature of event stream data, spatially and temporally. 
    \item We design a module called Bifurcated Folding Module (BFM) to encode the rich temporal information in the TAF tensor.
    \item We introduce a high-speed lightweight detector called Agile Event Detector (AED) along with a simple but effective data augmentation method. 
    \item We conduct experiments on two typical real-scene event camera object detection datasets. The experimental results show that compared with the state-of-the-art methods, our method has a far lower number of parameters and a much higher running speed. Moreover, our method retains competitive accuracy and superior motion robustness. 
\end{enumerate}

\section{Related works}

\begin{figure*}[htbp]
\centering
\includegraphics[width=\textwidth]{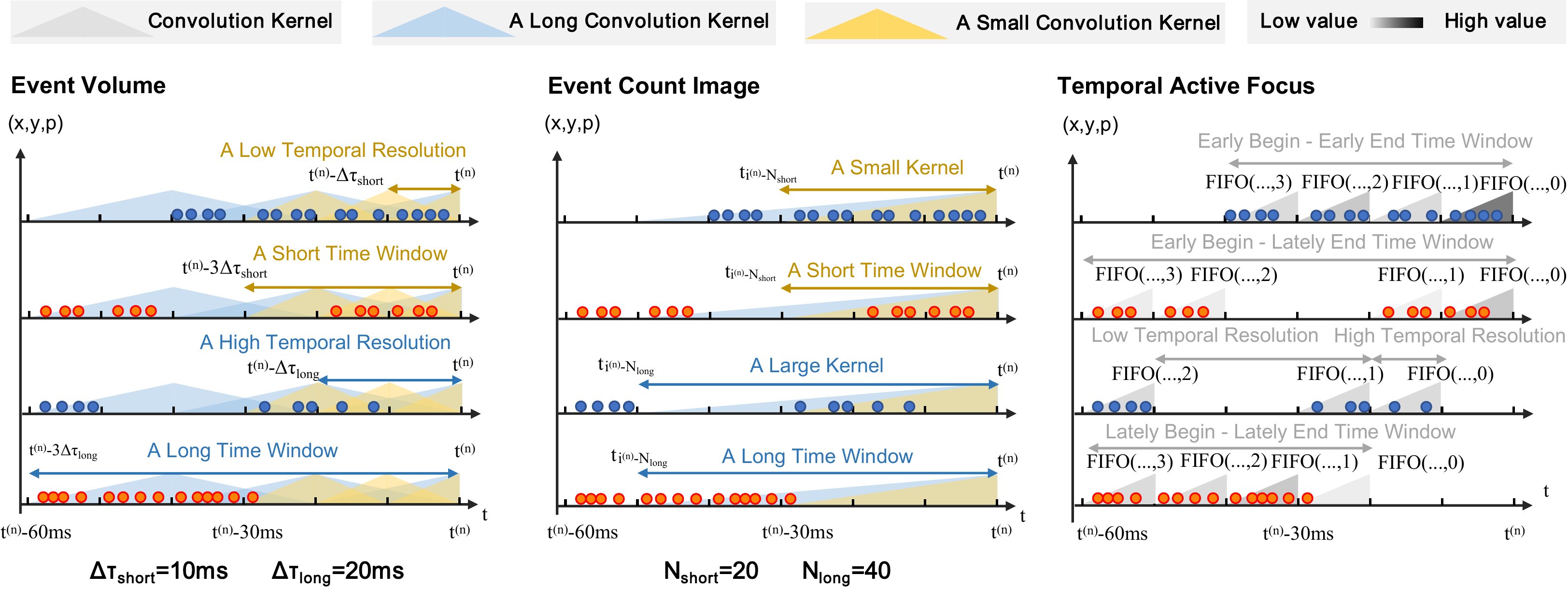}
\caption{Comparisons between the process of constructing Event Spike Tensors and the process of TAF at the $n^{th}$ detection. $t^{(n)}$ is the timestamp of the $n^{th}$ detection. $\Delta \tau$ is the sampling period. $B$ is the temporal dimension length. $N$ is the number of events to sample. }
\label{fig:2}
\end{figure*}

In this section, we present related works, including event representations and event-based object detection methods.

\subsection{Event representations}

Currently, representative approaches mainly process events in batches into tensor-like representations as the input of the object detection algorithms. Much research has demonstrated that the statistics of the event representation tensors overlap with those of natural images\cite{gehrig2019end,maqueda2018event,zhu2018ev}. The event representation methods are divided into two main categories: the Event Spike Tensor representation\cite{rebecq2017real,maqueda2018event,zhu2018ev,zhu2019unsupervised} and the representation of emulating spike neural networks\cite{benosman2013event,zhu2018ev,canniciasynchronous}. Event Spike Tensor is a data representation method based on discretizing the Event Measurement Field\cite{gehrig2019end}. The representation of emulating spike neural networks is to let the spatial locations of newly occurring events always have larger values in a tensor. 

There are hyper-parameters to be specified in all the existing event representation methods. The representative Event Spike Tensor requires hyper-parameters including the measurement function, the convolution kernel function, and timestamps for executing convolutions. There are also some works \cite{gehrig2019end,li2022asynchronous} that explore the possibility of learning the measurement function or the convolution kernel function directly from the raw event stream, but the timestamps for performing kernel convolutions are still hyper-parameters that need to be specified. The representation of emulating spike neural networks requires a global numerical transformation function as the hyper-parameter to keep the values in the tensor in a value range. The hyper-parameters determine the trade-off between the globally fixed time window length and temporal resolution. Therefore, the existing event representations cannot take into account objects with different velocities relative to the event camera. 

Inspired by a queue-based event representation method\cite{baldwin2022time}, we propose to explore its possibility in object detection tasks. We believe that its ability to asynchronously sample the event stream data at each spatial and polar position is significant to the object detection tasks. In our work, we propose a new kind of event stream representation method, solving the problem via asynchronously sampling the Event Measurement Field \cite{gehrig2019end} to adjust the time window length and temporal resolution on each spatial and polar position flexibly.

\subsection{Event stream object detectors}

Mainstream approaches use Convolutional Neural Networks (CNN) with a large number of parameters to build object detectors\cite{chen2018pseudo, hu2020learning, canniciasynchronous, messikommer2020event,yolov4_5d_2021_tim}. The main problems
with these methods are that they are not accurate, light, and fast enough for event stream data object detection tasks. Different from these methods, some approaches use recurrent convolutional neural network object detectors\cite{perot2020learning, li2022asynchronous} to improve the accuracy. However, the memory mechanisms used in the recurrent convolutional neural networks are not designed according to the characteristics of the event stream data. Therefore, those detectors are expensive to train and slow to run. In our work, we design a high-speed lightweight detector with competitive accuracy. Our detector is more suitable for event stream data object detection tasks.

\section{Problem Statements}\label{sec:theo}

A paradigm for the event-based object detection task was defined by Perot et al.\cite{perot2020learning}. It is briefly modified here to make it more precise and to facilitate the discussion later on: 

\begin{itemize}
    \item Event stream data: for a camera with the picture size height of $H$ and width of $W$, its event stream data is defined as $E:=\{e_i=(x_i,y_i,p_i,t_i)\}_{i\in \mathbb N}$, where $x_i\in \{0,1,... ,W-1\}$ and $y_i\in\{0,1,... ,H-1\}$ are the coordinates, $p_i\in\{0,1\}$ is the polarity, $t_i\in[0,T_{max})$ is the timestamp, $T_{max}$ is the maximum duration of the event stream record.
    \item Annotation: $B^*:=\{b^*_j=(x_j,y_j,w_j,h_j,l_j,t_j)\}_{j\in\mathbb N}$, where $x_j\in \mathbb R$ and $y_i\in \mathbb R$ are the coordinates of the upper left corner of a bounding box, $w_j\in \mathbb R^+$ is the width of the bounding box, $h_j\in \mathbb R^+$ is the height of the bounding box, $l_j\in\{0,... ,L\}$ is the category, $t_j$ is the timestamp. 
    \item Detection: $D(S(E^{(n)},t^{(n)}))$, where $D(.) $ is the detector, $S(.)$ is the event representation method, $n\in\{1,2,... \}$ means it is the $n^{th}$ detection on the event stream, $E^{(n)}:=\{e_i\}_{t_i\in[0,t^{(n)})}$ is the available event stream for the $n^{th}$ detection, $t^{(n)}\in[0,T_{max})$ denotes the timestamp of the $n^{th}$ detection. 
\end{itemize}

The event representation methods are divided into two main categories: the Event Spike Tensor representation\cite{rebecq2017real,maqueda2018event,zhu2018ev,zhu2019unsupervised} and the representation of emulating spike neural networks\cite{benosman2013event,zhu2018ev,canniciasynchronous}. 

The Event Spike Tensor methods can sample the Event Measurement Field\cite{gehrig2019end} with convolution kernel to represent rich information of the event stream. Among them, the Event Volume\cite{zhu2019unsupervised} is a typical Event Spike Tensor. It uses convolution kernel to sample events in the time window $t_i\in[t^{(n)}-B\Delta \tau,t^{(n)})$ at the $n^{th}$ detection, with a sampling resolution inversely proportional to the sampling period $\Delta \tau\in \mathbb R^+$ and a fixed temporal dimension length $B\in \mathbb Z^+$. The Event Count Image\cite{maqueda2018event,zhu2018ev}, on the other hand, uses a deformable convolution kernel to sample the recent $N\in \mathbb Z^+$ events. The kernel size and the time window covered can therefore be dynamically adjusted according to the frequency of events triggered within the recent event stream. However, the globally synchronized characteristic in these above representations has limitations. As illustrated in Fig. \ref{fig:2}, for spatial and polar locations that recently trigger events less frequently, a short global synchronized time window cannot aggregate enough information. On the contrary, for spatial and polar locations that recently trigger events more frequently, a long global synchronized time window leads to a decrease in the temporal resolution, making the convolution kernel cover excessive events, resulting in a loss of information. 

\begin{figure}[htbp]
\centering
\includegraphics[width=\linewidth]{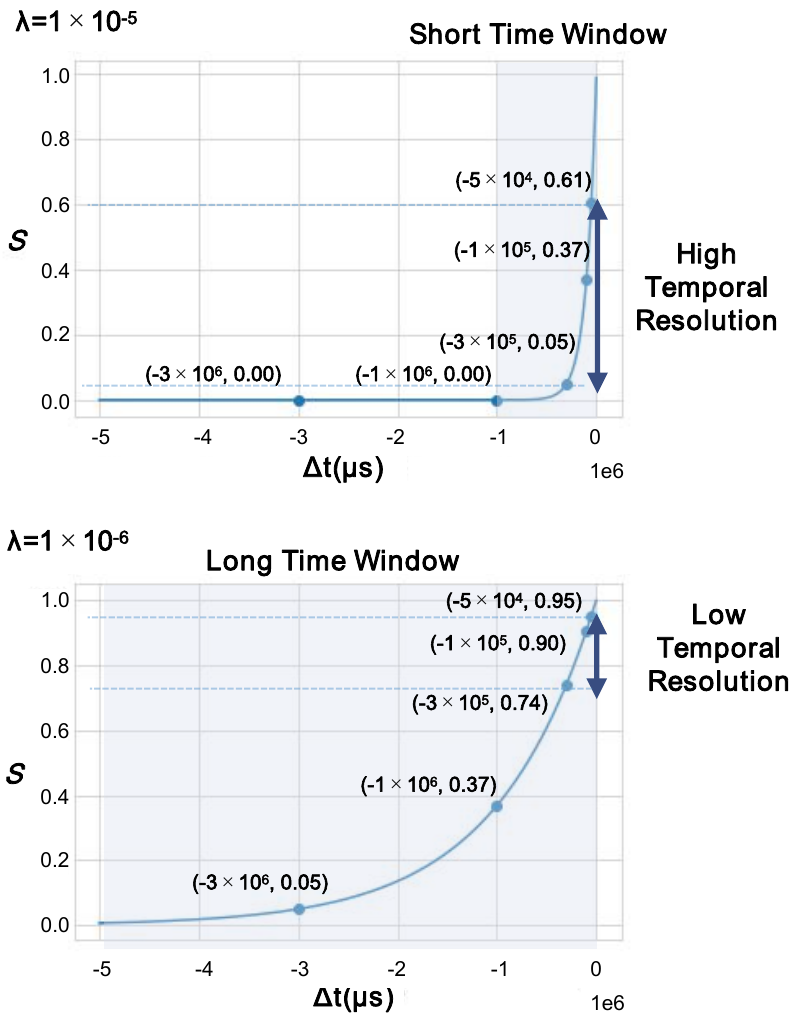}
\caption{The mapping between $\Delta t$ and the value in Surface of Active Events tensor under different $\lambda$. $\Delta t$ is the elapse from the timestamp of the $n^{th}$ detection $t^{(n)}$. $\lambda$ is a hyper-parameter. }
\label{fig:mapping}
\end{figure}

The methods of emulating spike neural networks do not apply a linear transformation to the Event Measurement Field, but the value of each spatial location is adjusted by discrete decisions during the event generation process. The generated representation tensors always satisfy that spatial locations of the recently generated events maintain larger values. Therefore, this kind of method supports flexible nonlinear temporal resolution adjustment. The surface of Active Events\cite{benosman2013event,zhu2018ev} is a typical example of the representation method of emulating spike neural networks. It gives a 2D snapshot of the latest timestamp of the events in the field of view. The elapse from the timestamp of the $n^{th}$ detection $\Delta t\in \mathbb R^-$ is then mapped to the value range $(0,1)$ with a numerical transformation $e^{\lambda \Delta t}$, where $\lambda\in \mathbb R^+$ is a hyper-parameter. Fig. \ref{fig:mapping} shows the mapping when $\lambda$ takes different values. It can be seen that under large $\lambda$, the mapping gives the newly triggered events a larger temporal resolution, while the information of events triggered further than a threshold from now will hardly be retained. Therefore, still, the Surface of Active Events can only consider events in a certain time window. The length of the time window can be increased by decreasing $\lambda$. However, on the other hand, the temporal resolution will be decreased. Therefore, since the hyper-parameter $\lambda$ is globally applied, the Surface of Active Events also faces the problem of globally synchronized characteristics like the Event Spike Tensor methods. 

\begin{figure*}[htbp]
\centering
\includegraphics[width=\textwidth]{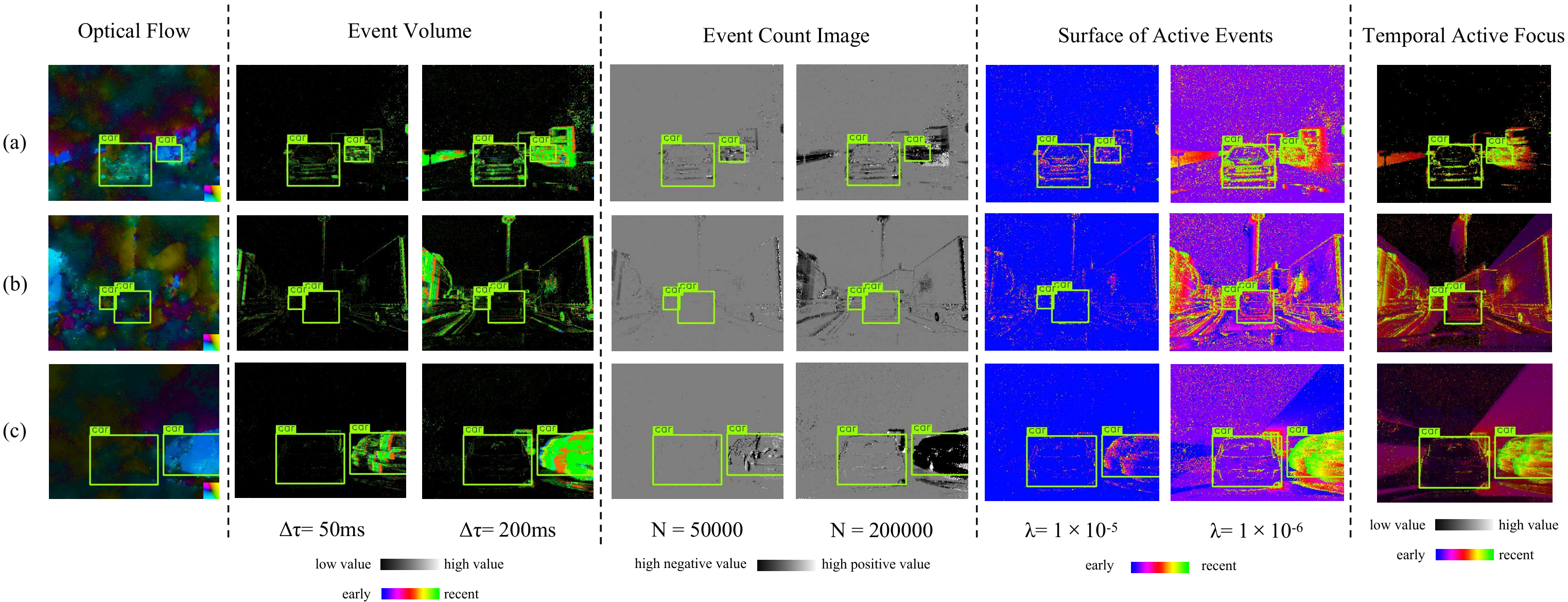}
\label{fig_ts}
\caption{Visualization of different event representation tensors with different hyper-parameters. Column 1 shows the TV-L1 optical flow plots of the scenes, which are computed by the method proposed by Nagata et al.\cite{nagata2021optical}. In the field of view, (a) is the case where only objects with fast motion relative to the event camera are present, (b) is the case where only objects with slow motion relative to the event camera are present, and (c) is the case where objects with different motion relative to event camera are present at the same time. }
\label{fig:visualization}
\end{figure*}

Fig.~\ref{fig:visualization} shows the visualization of different event representation tensors with different hyper-parameters. In summary, it is hard to find a hyper-parameter that takes into account all objects with different motion speeds relative to the camera in the field of view at the same time.

\section{Method}

This section introduces our core methods, including the event representation method Temporal Active Focus (TAF), the Bifurcated Folding Module (BFM) for fully extracting features from the TAF representation tensor, and the high-speed lightweight detector Agile Event Detector (AED). 

\subsection{Temporal Active Focus}

Tensors generated by the Temporal Active Focus method can be viewed as a dense version of the Event Spike Tensor. Considering a traditional Event Spike Tensor with variable length in temporal dimension to cover all event stream information by the time for detection, \textit{i.e.}, $B=\lceil \frac{t^{(n)}}{\Delta \tau}\rceil$. Due to the spatial and temporal sparsity of the event stream data\cite{messikommer2020event}, such a tensor is also a sparse tensor. Temporal Active Focus (TAF), by contrast, samples the first $K$ latest positions with a non-zero value in the temporal dimension at each spatial and polar position to form a dense tensor. 

The infinite-length sparse Event Spike Tensor is impractical to build in terms of time and storage cost. However, since the object detection on the event stream is performed at a certain frequency if the detection period is set to the sampling period $\Delta \tau$, the computational burden can be greatly reduced by using the FIFO queue to incrementally update the TAF tensor. The necessary condition is that the convolution kernel $k(\cdot) $ used for sampling the Event Measurement Field satisfies: 
\begin{equation}
\begin{aligned}
&\begin{cases}k(x,y,p,t)>0,\quad &t\in[0,k_{upper}]\\k(x,y,p,t)=0,\quad& Others\end{cases}\\s.t.\quad&k_{upper}\in [0,\infty)
\end{aligned}
\end{equation}

TAF maintains a per spatial and polar position specific FIFO queue $FIFO(x,y,p,k)$ with depth $K\in \mathbb Z^+$. The queues continuously receive non-zero samples from the event stream, thus generating dense tensor representations. The process of building the TAF tensor $\mathrm{S}\in \mathbb R^{2K\times H\times W}$ is shown in Algorithm \ref{alg1}. The convolution kernel $k(\cdot)$ and the measurement $f(\cdot)$ are essential components of Event Spike Tensor as introduced in \cite{gehrig2019end}. For the convolution kernel, we simply adopt the rectangular window function: \begin{equation}
    k(x,y,p,t)=\begin{cases}1,& t\in[0,\Delta \tau]\\0,&Others\end{cases}
\end{equation} In order not to lose the absolute position information on the temporal dimension, we use a measurement to calculate the average elapse from the events covered by the convolution kernel to $t^{(n)}$:\begin{equation}
f^{(n)}(x,y,p,t)=\frac{t^{(n)}-t}{\#\{E^{(n)}_{x,y,p}\}}
\end{equation} where $E^{(n)}_{x,y,p}=\{e_i\}_{t_i\in[t^{(n)}-k_{upper},t^{(n)}),x_i=x,y_i=y,p_i=p}$ and $\#(.)$ is the counting symbol.

At the $n$th detection, we have the average time elapses calculated as:\begin{equation}\begin{aligned}
&\Delta t^{(n)}(E,t,x,y,p):=\\&\sum_{e_i\in E}f(x_i,y_i,p_i,t_i,t^{(n)})k(x-x_i,y-y_i,p-p_i,t-t_i)\end{aligned}\end{equation} The non-zero values of the average time elapses will be pushed into the FIFO queues. Then at the $n+1$th detection, we have new values calculated and pushed into the queues, while old values are updated by: $\Delta t^{(n+1)}\longleftarrow\Delta t^{(n)}+\Delta \tau$.

\begin{figure}[htbp]
\centering
\includegraphics[width=0.8\linewidth]{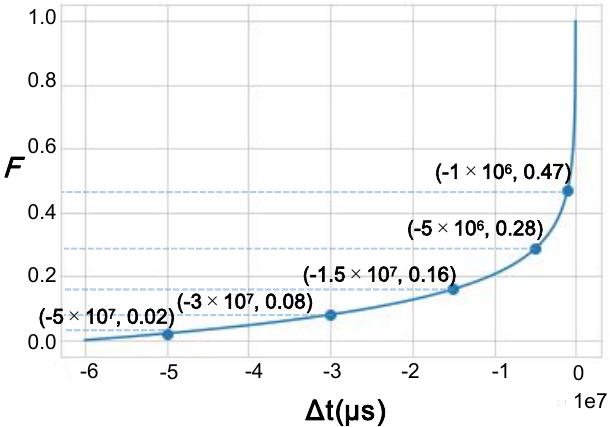}
\caption{The mapping between the elapse $\Delta t$ and the value of the measurement function $F(\cdot)$ when the maximum duration of the event stream record $T_{max}=6\times 10^7\mu s$. }
\label{fig:mapping log}
\end{figure}

To limit the range of $\Delta t$ values (in milliseconds), we apply another measurement function  based on the logarithmic transformation: \begin{equation}
    F(\Delta{t})=1-\frac{\ln(1+\Delta t\times10^{-4})}{\ln(1+T_{max}\times10^{-4})}
\end{equation} where $T_{max}$ is the maximum duration of the event stream (in milliseconds). Fig. \ref{fig:mapping log} shows that the values of $F(\cdot)$ on the whole$\Delta t\in[-T_{max},0)$ interval all reflect high resolutions. 

\begin{algorithm}[t]
	\caption{The process of building the TAF tensor when continuously performing object detection on the event data stream.}
	\label{alg1}
	\KwIn{$\mathcal{X}=\{0,1,...,W-1\}$,~$\mathcal{Y}=\{0,1,...,H-1\}$,\\$\mathcal{P}=\{0,1\}$,~$\mathcal{N}=\{1,2,...,\lceil\frac{T_{max}}{\Delta \tau}\rceil\}$,~$\mathcal{C}=\{0,1,...,2K-1\}$}
	\BlankLine
    \#Initialize FIFO queues
	
	\For{\rm{each} $x\in\mathcal{X}$,~$y\in\mathcal{Y}$,~$p\in\mathcal{P}$}{Initialize all values in $FIFO(x,y,p)$ with 0; }
	
    \#Detect on the event flow
	
	\For{\rm{each} $n\in \mathcal{N}$}{
        \#Perform convolution
	    
	    $E_{sub}^{(n)}:=\{e_i\}_{t_i\in[n\Delta\tau-k_{upper},n\Delta\tau)}$
	    
        \#Update FIFO queues
	    
	    \For{\rm{each} $x\in\mathcal{X}$,~$y\in\mathcal{Y}$,~$p\in\mathcal{P}$}{
	    
	    \If{$\Delta t^{(n)}(E^{(n)}_{sub},n\Delta \tau,x,y,p)>0$}{Add $\Delta t^{(n)}(E^{(n)}_{sub},n\Delta \tau,x,y,p)$ to $FIFO(x,y,p)$. }
     \For{\rm{each} $c\in\mathcal{C}$}{
	    Update $FIFO(x,y,p)[c]$ with new value}
		}
		
        \#Construct tensor
		
		\For{\rm{each} $x\in\mathcal{X}$,~$y\in\mathcal{Y}$,~$c\in\mathcal{C}$}{$S_{c,y,x}:=F\{FIFO(x,y,c-2\lfloor \frac{c}{2}\rfloor)[\lfloor \frac{c}{2}\rfloor]\}$}
		
		\KwOut{$\mathrm{S}:=\{S_{c,y,x}\}_{c\in\mathcal{C},y\in \mathcal{Y},x\in \mathcal{X}}$}  
	}
\end{algorithm}

\begin{figure*}
    \centering
\includegraphics[width=\linewidth]{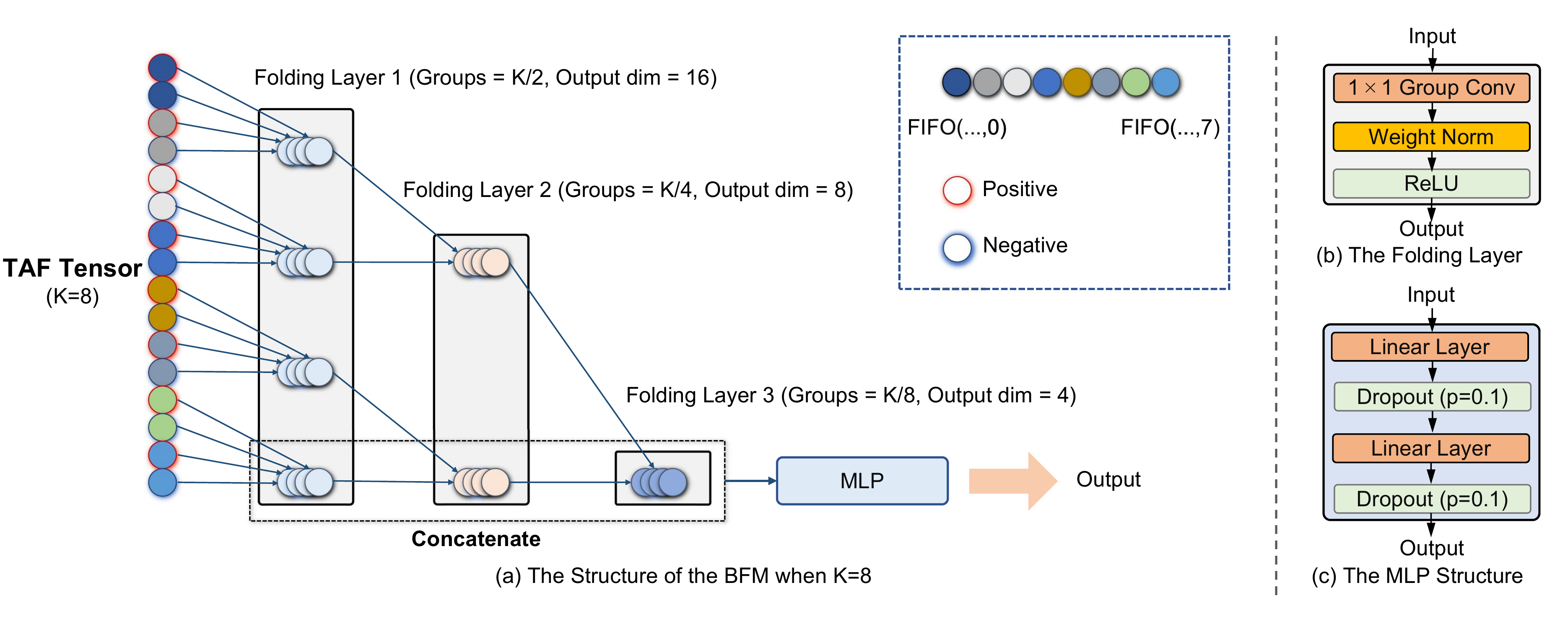}
\caption{(a) is the structure of the BFM module.  (b) and (c) are the structure of its components. $K$ is the length of the TAF queue.}
\label{fig:bfm}
\end{figure*}

Fig.~\ref{fig:2} visualizes the asynchronous characteristic of TAF, which shows that TAF can adapt the interval of applying convolutions asynchronously according to the event triggering frequency. Therefore, a larger temporal resolution is applied for periods with a higher event triggering frequency, while a smaller temporal resolution is applied otherwise. In addition, the time interval of event streams considered at each spatial and polar position starts and ends flexibly, which means that the time window can also be adjusted asynchronously. The positions and values in the temporal dimension will jointly encode information such as the elapse of the most recently triggered event at each position, the number of short-time triggered events, and the frequency of triggered events in a certain period, thus containing more information. Column 8 in Fig.~\ref{fig:visualization} shows the visualization of the information considered by Temporal Active Focus. It can be seen that the TAF tensor can consider equal high-resolution information for objects with different motion speeds.

\subsection{Bifurcated Folding Module}

The input layer of conventional object detectors is often a convolutional layer. The convolutional layer not only extracts information in the channel but also models spatial dependencies to extract spatial information. It works well when using the normal RGB image as the input since the semantics in channels are often weak. However, the TAF tensor has very rich temporal information encoded in the channels. Therefore, we design a module called the Bifurcated Folding Module (BFM) to extract semantic information in the temporal dimension of the TAF tensor before feeding it into the detector, which is point-wisely applied at each spatial location. 

The overall structure of the BFM is shown in Fig.~\ref{fig:bfm}(a). The design of the BFM follows two main concepts. First, instead of being fully connected to the temporal dimension, it will gradually aggregate values at adjacent temporal positions to prevent over-fitting. Second, it will assign greater importance to recent information as it has a greater temporal resolution. 

Following the first concept, we design the Folding Layer. The structure of the Folding Layer is shown in Fig.~\ref{fig:bfm}(b). The Folding Layer gradually reduces the number of input channels using the $1\times 1$ depth-wise convolution. It actually models the local connection of adjacent temporal positions. The structure of the Folding Layer and its function is similar to Temporal Convolutional Network (TCN)\cite{lea2017temporal}, so we imitate the design of TCN with Weight Normalization and ReLU activation followed. 

Following the second concept, we are inspired by the Cross-Stage-Partial-connections (CSP)\cite{bochkovskiy2020yolov4} mechanism. After each temporal aggregation, we make a slicing in the channel dimension to get the channels encoding the temporally latest information and finally connect all sliced channels. This operation also acts as the residual connection, making the module easier to converge during training\cite{he2016deep}. We finally fuse the information encoded in output channels with a Multi-Layer Perceptron (MLP). The structure of the MLP is shown in Fig.~\ref{fig:bfm}(c). 

Some alternative approaches such as Temporal Convolutional Network (TCN)~\cite{lea2017temporal}, Recurrent Neural Networks (RNNs)~\cite{jordan1997serial, hochreiter1997long}, and Transformer-based models~\cite{vaswani2017attention} can be used to extract temporal information from TAF tensors. However, the BFM model has the following advantages. Firstly, BFM is more lightweight and computationally efficient, which makes it more suitable for event data that has a high temporal resolution. Secondly, BFM is more closely combined with TAF event representation since it is designed to model the local connection of adjacent temporal positions rather than long-range temporal dependencies. Moreover, it also explicitly assigns great importance to the temporal latest information in the TAF data since they have a higher temporal resolution.

\subsection{Agile Event Detector}

\begin{figure}[htbp]
\centering
\includegraphics[width=\linewidth]{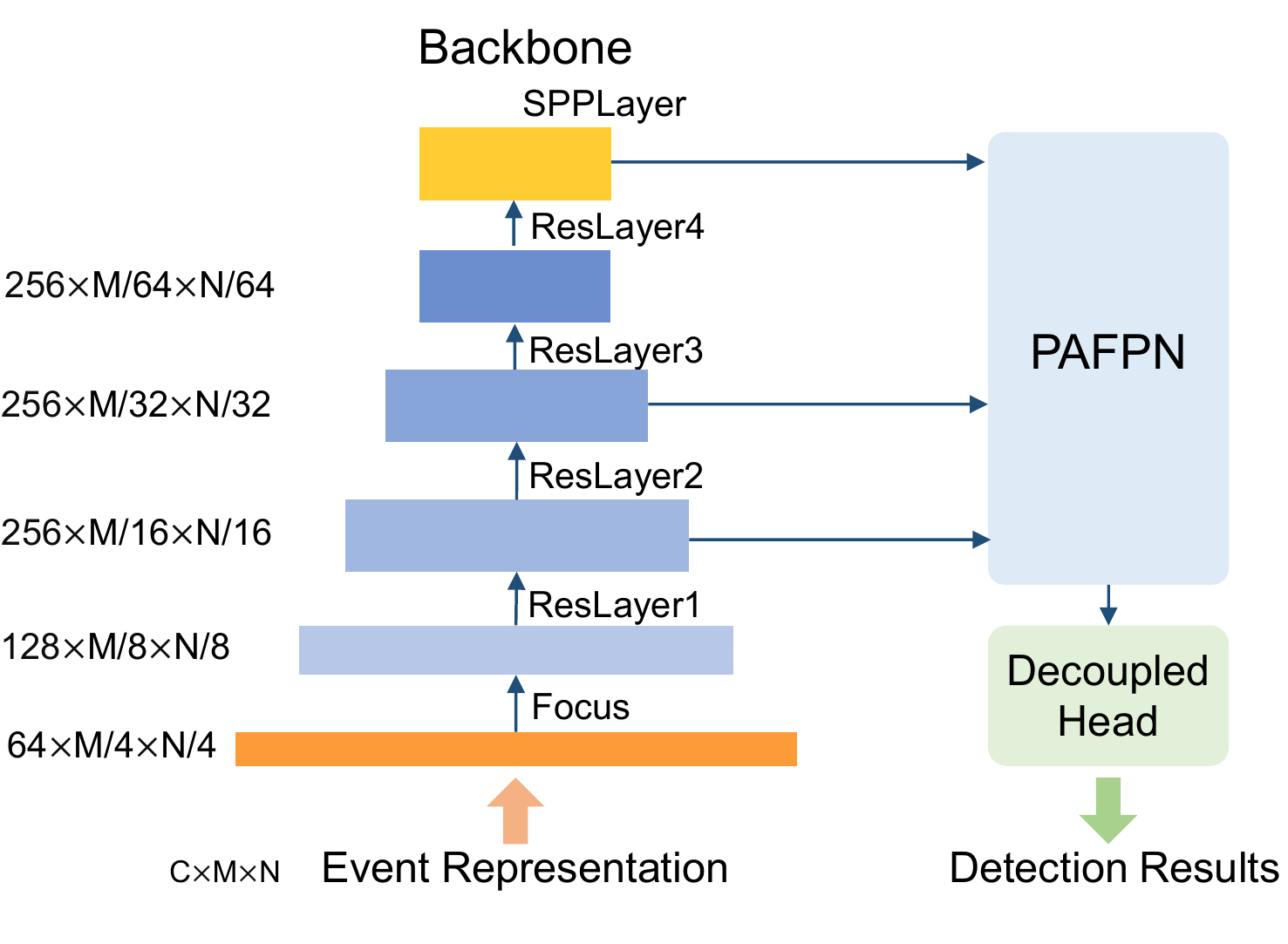}
\caption{The structure of the AED model. It contains a backbone network adopted from the Darknet21\cite{redmon2018yolov3}. It also contains an FPN  and a detection head the same as the implementation in YOLOX\cite{ge2021yolox}. $(C\times M\times N)$ is the size of the input event representation tensor. }
\label{fig:aed structure}
\end{figure}

In the field of generic object detection, YOLO series~\cite{chen2018pseudo,hu2020learning,canniciasynchronous} are featured by the outstanding trade-off between accuracy and efficiency. The YOLO series has the advantage of high speed over other approaches such as Faster R-CNN~\cite{ren2015faster}, Mask R-CNN~\cite{he2017mask}, RetinaNet~\cite{Focal-Loss_2017_ICCV}, and DETR~\cite{detr_2020_eccv}, while maintaining competitive detection accuracy. Hence, we customize an Agile Event Detector (AED) based on the YOLOX model \cite{ge2021yolox}, which is more lightweight and yields higher speed compared to the YOLOX baseline. The general structure of the AED model is shown in Fig.~\ref{fig:aed structure}. 

Different from the YOLOX's CSPDarknet\cite{bochkovskiy2020yolov4} backbone, AED utilizes a backbone network adapted from the Darknet21 used in YOLOv3\cite{redmon2018yolov3}. Since there is no channel connection operation, Darknet21 runs faster than CSPDarknet. In addition, we increase the channel number of feature maps at an early stage since the event representation tensors encode rich semantics in the channels. The Feature Pyramid Network (FPN) and the detection head are modified from the YOLOX as shown in Fig.~\ref{fig:aed structure}. 

To improve the generalization capability of the lightweight detector, we also propose a simple but effective data augmentation method for event stream representation tensor data: 

\begin{enumerate}
    \item Random Flipping: During training, each event representation tensor $S$ has $p_1$ probability to be flipped horizontally, \textit{i.e.}, $S^*_{c,y,x}=S_{c,y,W-x-1}$. 
    \item Random Cropping and Resizing: During training, each event representation tensor $\mathrm S\in \mathbb{R}^{C\times H\times W}$ has $p_2$ probability to be resampled to $ C\times \lfloor\alpha H\rfloor\times \lfloor\alpha W\rfloor$ using nearest neighbor interpolation, where $\alpha\in[1,\infty)$, and then randomly cropped back to $C\times H\times W$, \textit{i.e.} $S^*_{c,y,x}=S_{c,y^*:y^*+H,x^*:x^*+W}$, where $y^*\sim U(0,\lfloor\alpha H\rfloor-H),x^*\sim U( 0,\lfloor\alpha W\rfloor-W)$, and $U(a,b)$ is the uniform distribution on $[a,b], a, b\in\mathbb R$. 
\end{enumerate}

\section{Experimental Results}

\begin{figure*}[htbp]
\centering
\includegraphics[width=\linewidth]{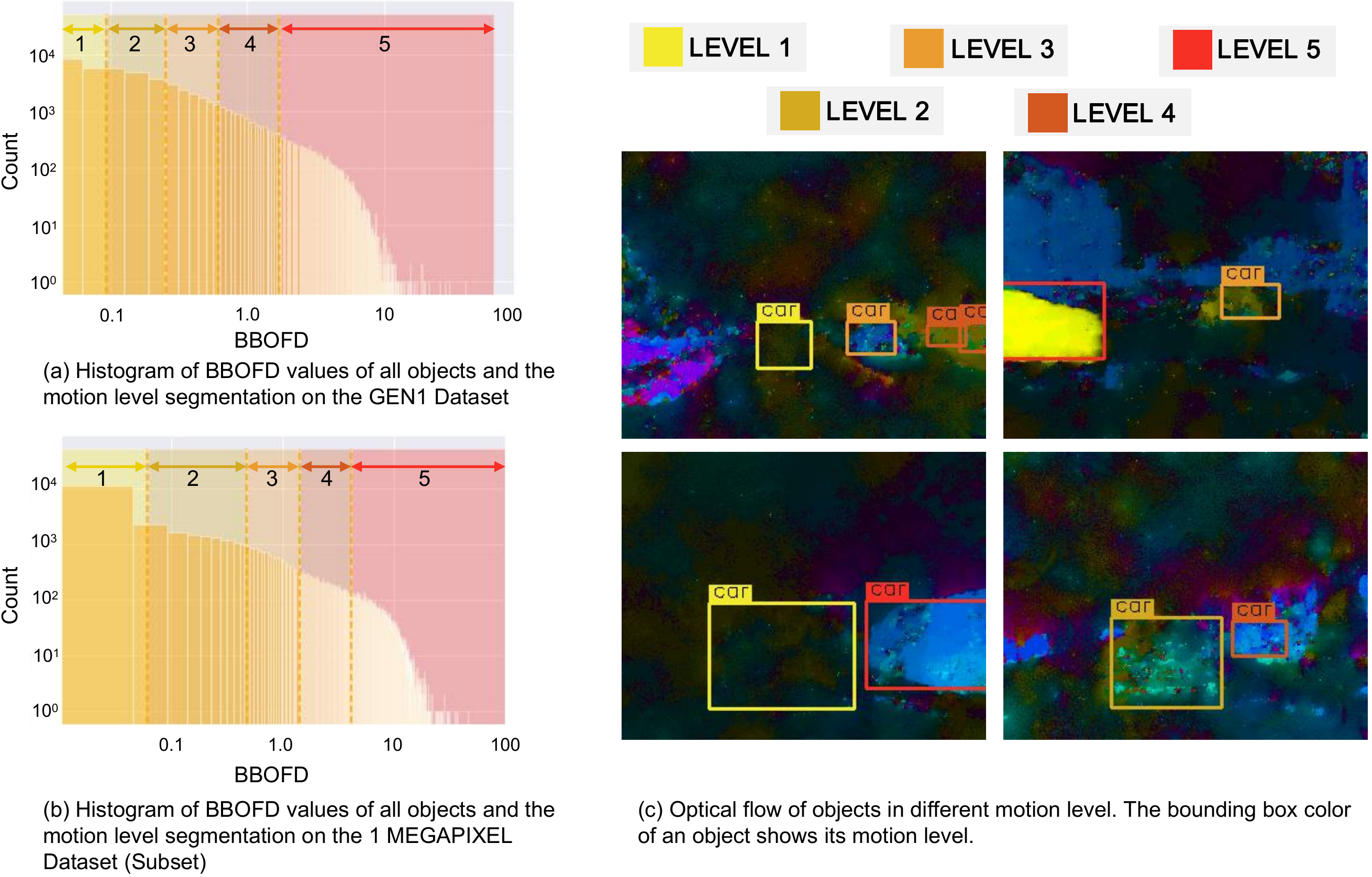}
\caption{The visualization of data segmentation and examples of objects under different motion levels.}
\label{fig:segmentation2}
\end{figure*}

In this section, we will present our experimental settings, including details of the dataset used for the experiment, the selection of hyper-parameters, and the evaluation method. Then we will compare our method with state-of-the-art methods. Finally, we will illustrate the effectiveness of each component of our method through ablation studies. 

\subsection{Experiment Settings}

\subsubsection{Dataset}

We choose the Prophesee GEN1 Automotive Detection Dataset (GEN1 Dataset)\cite{de2020large} and the Prophesee 1 MEGAPIXEL Automotive Detection Dataset (1 MEGAPIXEL Dataset)\cite{perot2020learning} to conduct the experiments. The 1 MEGAPIXEL Dataset and the GEN1 Dataset are two typical real-scene event camera object detection datasets. Each video stream is up to 60 seconds, captured with the event camera and stored as the event stream. The timestamps are in microsecond resolution, therefore $T_{max}=6\times 10^7\mu s$. 

To improve the efficiency of performance validation, our experiments are conducted on the complete GEN1 Dataset and the downsampled 1 MEGAPIXEL Dataset (1 MEGAPIXEL Dataset (Subset)) by reducing the annotation frequency to 1 Hz while preserving the complete event stream data. This downsampling approach preserves as much diversity of scenes and objects as possible. However, since the number of samples used for training is only 1/60 of the original dataset, the performance of the trained model may be degraded. 

\subsubsection{Implementation Details}

Referring to the setting in YOLOX\cite{ge2021yolox}, all methods based on YOLOX and AED models are trained using Adam optimizer\cite{kingma2014adam} and cosine learning rate with 5 preheat epochs. In the preheat epochs, the learning rate will grow from 0 to $2.1\times 10^{-4}\times Batchsize$, and then decrease. The Batch size is taken as 30 on GEN1 Dataset and 16 on the 1 MEGAPIXEL Dataset (Subset). We use the best training model on the validation dataset and apply it to the testing dataset to report the final performance, which is a widely used method in related works\cite{perot2020learning, li2022asynchronous}. For the data augmentation, we take $p_1=0.5$, $p_2=0.5$, and $\alpha=1.5$. For all Event Volume representation tensors, we take $B=5$, which is commonly used in existing work\cite{perot2020learning}. For the TAF tensor, we set $\Delta\tau=10ms$ on both datasets since it approximates the run time of the method. All methods in our work were trained on a server with a single GeForce RTX 3090 GPU and an 8-core Intel(R) Core(TM) i7-7820X CPU @ 3.60GHz. We test all methods on a server with a single Titan Xp GPU and a 16-core Intel(R) Xeon(R) CPU E5-2620 v4 @ 2.10GHz. 

\subsubsection{Evaluation Method}

The performance metrics include accuracy, running time, and the number of parameters. The accuracy evaluation protocol is consistent with the evaluation protocol provided by the 1 MEGAPIXEL Dataset \cite{perot2020learning}. In the protocol, the Mean Average Precision (mAP)~\cite{everingham2009pascal} is used as the accuracy metric, which is a commonly used metric for object detection models. mAP~(IoU@[0.5:0.05:0.95]) is used for evaluation, where the mAP is computed by the mean over a range of IoU thresholds from 0.5 to 0.95 with a step size of 0.05. For algorithms that use the TAF tensor as input, we set the detection period equal to $\Delta\tau$ and the tolerance to $\Delta\tau/2$; for algorithms using other event representation methods, we specify detections at timestamps where the annotations appear. 

To evaluate the motion robustness of methods, we measure the motion speed of objects in the dataset by computing optical flow, to classify them into 5 different motion levels. To be specific, we compute the TV-L1 dense optical flow using the method proposed by Nagata et al. \cite{nagata2021optical} for all timestamps where the annotations are present. The difference between adjacent timestamps is set to $50ms$. For timestamp $t_j$ with an annotation $b^*_j$ presents, for a camera with the picture size of $H$ in height and $W$ in width, let the estimated horizontal optical flow be $u_{t_j}(x,y)$ and the vertical optical flow be $v_{t_j}(x,y)$, then we have the optical flow intensity denoted as:  \begin{center}\begin{equation}\begin{aligned}
    &||V(x,y)||_{t_j}=\sqrt{u^2_{t_j}(x,y)+v^2_{t_j}(x,y)}\\\forall\quad&x\in\{0,1,... ,W-1\},y\in\{0,1,... ,H-1\}
\end{aligned}\end{equation}\end{center} Based on the optical flow intensity, we can define a metric called Bounding Box Optical Flow Density (BBOFD) for each annotation $b^*_j$: \begin{center}\begin{equation}||V(x,y)||_j=\frac{\sum_{x=x_j}^{x_j+w_j-1}\sum_{y=y_j-h_j+1}^{y_j}||V(x,y)||_{t_j}}{w_j\times h_j}\end{equation}\end{center} For the object corresponding to the annotation $b^ *_j$, BBOFD can be an effective metric for its motion speed relative to the event camera at $t_j$. The larger the $||V(x,y)||_j$, the faster the object. Of all the annotations in each of the two datasets, we set 5 intervals for the BBOFD at each 20\% quantile. The intervals can classify the annotations and detection results into 5 classes according to their corresponding objects' speed relative to the camera. We call the 5 classes the 5 motion levels, which are numbered from 1-5, indicating the speed from low to high. We show the visualization of data segmentation and examples of objects under different motion levels in Fig. \ref{fig:segmentation2}. 

To make the metric accurate, we filtered out all overlapping bounding boxes and cropped all bounding boxes into the camera picture size. We then evaluate the detection results under each of the five classes. 

\subsection{Comparison with the State-of-the-art}

\begin{table*}[htbp]
    \caption{Performance comparison with the state-of-the-art methods. The bold is the best result in each group of comparisons. }
    \centering
    \begin{tabular}{cccc|cc|cc|c}
    \toprule[1.5pt]
        \multirow{2}{*}{Method} & \multirow{2}{*}{Event Representation} & \multirow{2}{*}{Detector} & \multirow{2}{*}{Memory} & \multicolumn{2}{c|}{GEN1 Dataset} & \multicolumn{2}{c|}{\thead{1 MEGAPIXEL\\Dataset (Subset)}} & \multirow{2}{*}{Params(M)} \\ \cline{5-8}
         &  &  &   & mAP & \thead{Inference\\Time(ms)} & mAP & \thead{Inference\\Time(ms)} &  \\ \hline
        Chen et al.\cite{chen2018pseudo} & Event Count Image\cite{rebecq2017real} & YOLO\cite{redmon2016you} & - & 0.322 & 21.47 & - & - & 45.3\\  Jiang et al.\cite{jiang2019mixed} & Event Count Image\cite{rebecq2017real} & YOLOv3\cite{redmon2018yolov3} & - & 0.326 & 22.34 & 0.207 & 20.81 & 61.5 \\  JDF-events \cite{li2019event} & 2 Polarities Event Count Image\cite{rebecq2017real} & YOLOv3\cite{redmon2018yolov3} & - & 0.332 & 22.34 & 0.224 & 20.81 & 61.5 \\  NGA-events \cite{hu2020learning} & Event Volume\cite{ zhu2019unsupervised} & YOLOv3\cite{redmon2018yolov3} & - & 0.359 & 26.11 & 0.248 & 21.23 & 61.5 \\  Sparse-conv \cite{messikommer2020event} & Raw Events & YOLO\cite{redmon2016you} & - & 0.145 & - & - & - & -\\  RED \cite{perot2020learning} & Event Volume\cite{ zhu2019unsupervised} & SSD\cite{liu2016ssd} & ConvLSTM\cite{shi2015convolutional} & 0.400 & - & - & - & 24.1 \\  ASTMNet \cite{li2022asynchronous} & Raw Events & SSD\cite{liu2016ssd} & Rec-conv\cite{li2022asynchronous} & \textbf{0.467} & 35.61 & - & - & 39.6 \\  \hline Our baseline & Event Volume\cite{ zhu2019unsupervised} & YOLOX\cite{ge2021yolox} & - & 0.350 & 13.19 & 0.213 & 16.16 & \textbf{14.4} \\ Ours & Temporal Active Focus & AED & - & 0.454 & \textbf{11.98} & \textbf{0.344} & \textbf{13.36} & 14.8 \\ \bottomrule[1.5pt]
    \end{tabular}
    \label{tab:sota}
\end{table*}

We compare our method with the state-of-the-art methods\cite{chen2018pseudo, jiang2019mixed, li2019event, hu2020learning, perot2020learning, li2022asynchronous}. We first compare our overall method with other existing works in terms of accuracy, model inference speed, and model parameters. Then, for the event representation methods, we conduct a separate comparison experiment, using AED as the detector, comparing the accuracy and speed of TAF with other methods. 

\subsubsection{Comparison with the State-of-the-art methods }

TABLE \ref{tab:sota} shows the results of comparing our method with the state-of-the-art methods\cite{chen2018pseudo, jiang2019mixed}. It can be seen that among the methods using feed-forward detectors without the memory mechanism \cite{chen2018pseudo, jiang2019mixed, li2019event, hu2020learning, messikommer2020event}, our method achieves the best performance in terms of accuracy and speed at the same time. Compared with the NGA-events method\cite{hu2020learning}, our method needs only 24.1\% of parameters but achieves 9.5 mAP points improvement and 54.1\% model inference time reduction on the GEN1 Dataset. Our method also achieves 9.6 mAP points improvement and 37.1\% inference time reduction on the 1 MEGAPIXEL Dataset (Subset). ASTMNet\cite{li2022asynchronous} is currently the most advanced method on GEN1 Dataset. Comparing our method with ASTMNet, the mAP is 1.3 points lower on GEN1 Dataset, but the model inference time reduces by 66.4\%, and the number of model parameters is only 37.4\%. It can be seen that our method is competitive among existing methods when jointly considering the accuracy, speed, and model size metrics. 

\subsubsection{Comparison with the State-of-the-art event representation methods}

\begin{table*}[htbp]
    \caption{Comparison of accuracy between different event representation methods with different parameters at 5 motion levels on GEN1 Dataset. The bold is the best result in each group of comparisons. }
    \centering
    \begin{tabular}{cccc|ccccccc}
    \toprule[1.5pt]
        Method & $\Delta\tau(ms)$ & $N$ & $\lambda$ & \thead{Lv1\\mAP} & \thead{Lv2\\mAP} & \thead{Lv3\\mAP} & \thead{Lv4\\mAP} & \thead{Lv5\\mAP} & \thead{Overall\\mAP} & \thead{Representation\\Time(ms)} \\ \hline
        \multirow{3}{*}{Event Volume\cite{ zhu2019unsupervised}} & 50 & \multirow{3}{*}{-} & \multirow{3}{*}{-} & 0.172 & 0.414 & 0.455 & 0.451 & 0.397 & 0.426 & 2.60 \\  & 100 &  & & 0.198 & 0.428 & 0.453 & 0.441 & 0.382 & 0.424 & 2.80 \\   & 200 &  &  & 0.218 & 0.434 & 0.459 & 0.448 & 0.374 & 0.422 & 3.27 \\  \multirow{3}{*}{Event Count Image\cite{maqueda2018event,zhu2018ev}} & \multirow{3}{*}{-} & $5\times 10^4$  & \multirow{3}{*}{-} & 0.204 & 0.371 & 0.408 & 0.411 & 0.362 & 0.386 & \textbf{0.90}\\  & & $1\times 10^5$ & & 0.211 & 0.377 & 0.405 & 0.395 & 0.343 & 0.376 & 1.01\\   & & $2\times 10^5$ & & 0.228 &0.388 & 0.394 & 0.369 & 0.312  & 0.368 & 1.19 \\ \multirow{3}{*}{Surface of Active Events\cite{ benosman2013event,zhu2018ev}} & \multirow{3}{*}{-} & \multirow{3}{*}{-} & $1\times 10^{-5}$ & 0.199 & 0.403 & 0.425 & 0.421 & 0.367 & 0.400 & 0.99\\  &  &  & $2.5\times 10^{-6}$ & 0.216 & 0.429 & 0.431 & 0.403 & 0.353 & 0.404 & 1.01\\   &  &  & $1\times 10^{-6}$ & 0.228 & 0.420 & 0.428 & 0.400 & 0.349 & 0.403 & 1.06\\  \hline  Temporal Active Focus  & 10 & - & - & \textbf{0.282} & \textbf{0.461} & \textbf{0.476} & \textbf{0.459} & \textbf{0.407} &\textbf{0.454} & 1.43\\ \bottomrule[1.5pt]
    \end{tabular}
    \label{tab:5motion Gen1}
\end{table*}

\begin{table*}[htbp]
    \caption{Comparison of accuracy between different event representation methods with different parameters at 5 motion levels on 1 MEGAPIXEL Dataset (Subset). The bold is the best result in each group of comparisons. }
    \centering
    \begin{tabular}{cccc|ccccccc}
    \toprule[1.5pt]
        Method & $\Delta\tau(ms)$ & $N$ & $\lambda$ & \thead{Lv1\\mAP} & \thead{Lv2\\mAP} & \thead{Lv3\\mAP} & \thead{Lv4\\mAP} & \thead{Lv5\\mAP} & \thead{Overall\\mAP} & \thead{Representation\\Time(ms)} \\ \hline
        \multirow{3}{*}{Event Volume\cite{ zhu2019unsupervised}} & 50 & \multirow{3}{*}{-} & \multirow{3}{*}{-} & 0.038 & 0.219 & 0.224 & \textbf{0.318} & \textbf{0.314} & 0.299 & 7.80 \\  & 100 &  & & 0.035 & 0.232 & 0.325 & 0.315 & 0.301 & 0.305 & 11.14 \\   & 200 &  &  & 0.062 & 0.239 & 0.309 & 0.281 & 0.276 & 0.290 & 14.20 \\  \multirow{3}{*}{Event Count Image\cite{maqueda2018event,zhu2018ev}} & \multirow{3}{*}{-} & $4\times 10^5$  & \multirow{3}{*}{-} & 0.111 & 0.209 & 0.269 & 0.257 & 0.271 & 0.276 & \textbf{1.31}\\  & & $8\times 10^5$ & & 0.105 & 0.212 & 0.270 & 0.253 & 0.254 & 0.268 & 1.34\\   & & $1.2\times 10^6$ & & 0.131 & 0.233 & 0.258 & 0.240  & 0.260 & 0.278 & 1.49 \\ \multirow{3}{*}{Surface of Active Events\cite{ benosman2013event,zhu2018ev}} & \multirow{3}{*}{-} & \multirow{3}{*}{-} & $1\times 10^{-5}$ & 0.043 & 0.231 & 0.315 & 0.287 & 0.286 & 0.294 & 3.71\\  &  &  & $2.5\times 10^{-6}$ & 0.090 & 0.234 & 0.297 & 0.274 & 0.288 & 0.296 & 3.20\\   &  &  & $1\times 10^{-6}$ & 0.143 & 0.262 & 0.300 & 0.272  & 0.296  & 0.303 & 3.32\\  \hline  Temporal Active Focus  & 10 & - & - & \textbf{0.262} & \textbf{0.312} & \textbf{0.339} & 0.308 & \textbf{0.314} &\textbf{0.344} & 2.83\\ \bottomrule[1.5pt]
    \end{tabular}
    \label{tab:5motion megapix}
\end{table*}

Based on the AED with data augmentation, we compare our TAF representation with three state-of-the-art competitors, namely, Event Volume\cite{ zhu2019unsupervised}, Event Count Image\cite{maqueda2018event,zhu2018ev}, and Surface of Active Events\cite{benosman2013event,zhu2018ev}. We set three sets of hyper-parameters for each of the three event representation methods, resulting in different global synchronized time windows and temporal resolutions. 

TABLE \ref{tab:5motion Gen1} and \ref{tab:5motion megapix} show the comparison of accuracy between different event representation methods with different parameters at 5 motion levels in the GEN1 Dataset and the 1 MEGAPIXEL Dataset (Subset) respectively. Though with different settings, the accuracy of different methods varies at each motion level, different methods achieve the highest accuracy on motion level 3 in general. On the one hand, as the motion level goes lower, the accuracy reduces, since the events are triggered less frequently and event representation methods cannot aggregate enough information. On the other hand, as the motion level goes higher, the accuracy also degrades, which results from the fact that fast-moving objects generate excessive events, causing motion blur in the event representation. It can also be found that in general the accuracy on the motion level 1 is much lower than that on level 5. This indicates that in the event data object detection task, the accuracy degradation caused by insufficient information is more severe than that caused by motion blur. It can be seen that our TAF method is highly competitive in terms of both accuracy and speed. When using Event Count Image, Surface of Active Events, and Event Volume, longer time windows should be used to improve the detection accuracy for objects with low motion levels. On the contrary, shorter time windows should be used for objects with high motion levels. This demonstrates the trade-off that exists in these event representation methods. By contrast, the improvement in the time window and temporal resolution makes the TAF method achieves high detection accuracy at all motion levels in a hyper-parameter-free manner. Especially when detecting low-motion level objects, using the TAF method can achieve quite significant accuracy improvements. 

\begin{figure*}[htbp]
\centering
\includegraphics[width=\textwidth]{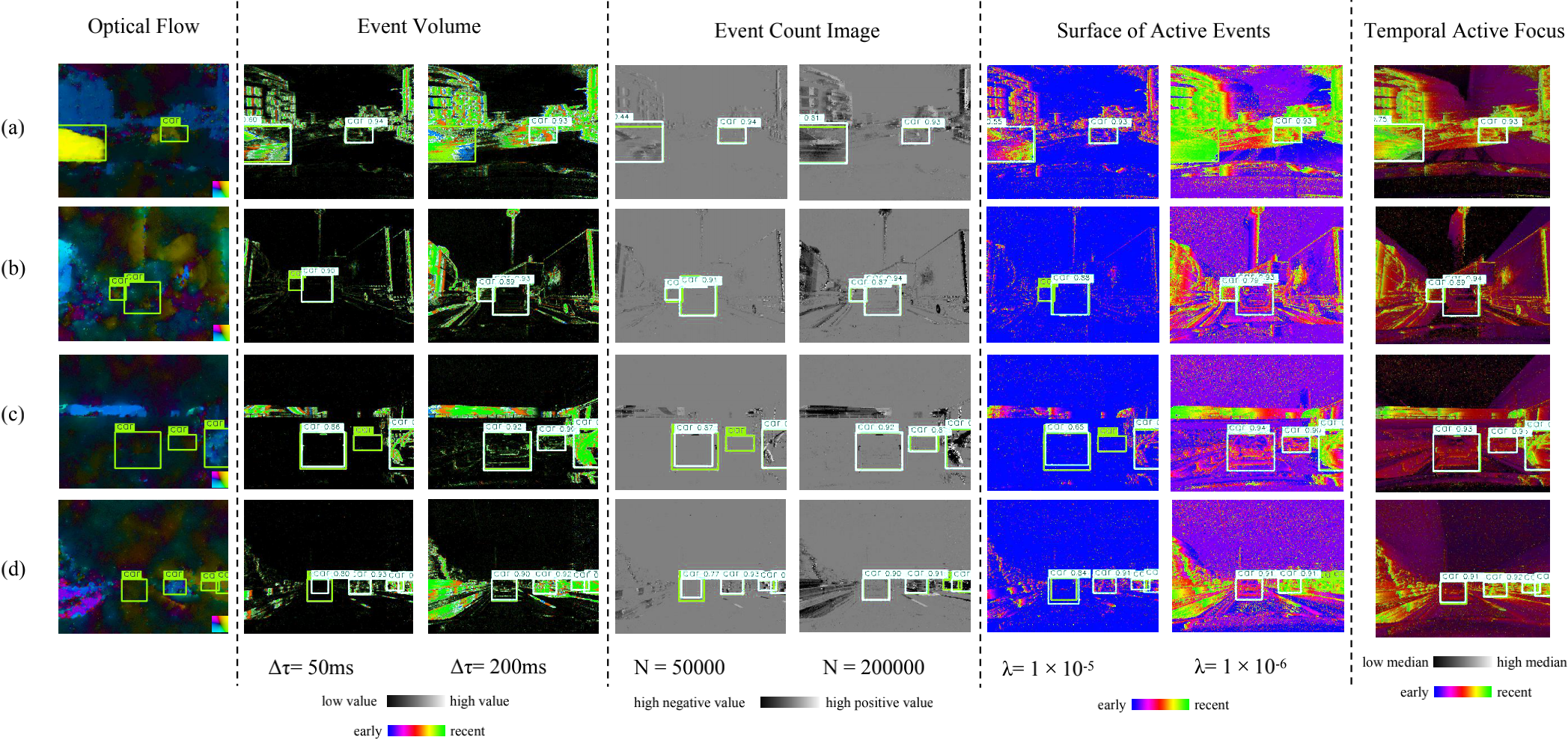}
\caption{Qualitative analysis results. The green bounding boxes indicate the annotations, while the white bounding boxes indicate the detection results.}
\label{fig:qualitative}
\end{figure*}

\begin{table*}[htbp]
    \caption{The ablation of each component in our method. The bold is the best result in each group of comparisons. }
    \centering
    \begin{tabular}{ccc|cc|cc|c}
    \toprule[1.5pt]
        \multirow{2}{*}{Detector} & \multirow{2}{*}{Data Augmentation} & \multirow{2}{*}{Event Representation} &  \multicolumn{2}{c|}{GEN1 Dataset} & \multicolumn{2}{c|}{\thead{1 MEGAPIXEL\\Dataset (Subset)}} & \multirow{2}{*}{Params(M)} \\ \cline{4-7}
         &  &  &  mAP & Runtime(ms) & mAP & Runtime(ms) &  \\ \hline
        YOLOX\cite{ge2021yolox} & - & Event Volume\cite{ zhu2019unsupervised} & 0.350 & 15.79 & 0.213 & 23.96 & \textbf{14.4}\\  
        YOLOX\cite{ge2021yolox} & \checkmark & Event Volume\cite{ zhu2019unsupervised} & 0.410 & 15.79 & 0.269 & 23.96 & \textbf{14.4}\\
        AED & \checkmark & Event Volume\cite{ zhu2019unsupervised} & 0.426 & 14.18 & 0.299 & 21.35 & 14.8\\
        AED & \checkmark & TAF  & \textbf{0.454} & \textbf{13.41} & \textbf{0.344} & \textbf{16.19} & 14.8\\ \bottomrule[1.5pt]
    \end{tabular}
    \label{tab:ablation}
\end{table*}

\begin{table*}[htbp]
    \caption{The effects of inserting TAF to SOTA methods. The bold is the best result in each group of comparisons.}
    \centering
    \begin{tabular}{cc|cc|cc|c}
    \toprule[1.5pt]
        \multirow{2}{*}{Detector} & \multirow{2}{*}{Event Representation} &  \multicolumn{2}{c|}{GEN1 Dataset} & \multicolumn{2}{c|}{\thead{1 MEGAPIXEL\\Dataset (Subset)}} & \multirow{2}{*}{Params(M)} \\ \cline{3-6}
         &  &  mAP & Runtime(ms) & mAP & Runtime(ms) &  \\ \hline
        \multirow{2}{*}{YOLOv3\cite{redmon2018yolov3}} & Event Volume\cite{ zhu2019unsupervised} & 0.359 & 28.71 & 0.248 & 29.03 & \textbf{61.5}\\  
          & TAF & \textbf{0.381} & \textbf{27.54} & \textbf{0.278} & \textbf{24.06} & 61.6\\\hline\multirow{2}{*}{YOLOX\cite{ge2021yolox}} & Event Volume\cite{ zhu2019unsupervised} & 0.410 & 15.79 & 0.269 & 23.96 & \textbf{14.4}\\  & TAF & \textbf{0.436} & \textbf{14.62} & \textbf{0.314} & \textbf{18.99} & 14.4\\ \hline
        \multirow{2}{*}{AED}  & Event Volume\cite{ zhu2019unsupervised} & 0.426 & 14.18 & 0.299 & 21.35 & \textbf{14.8}\\
         & TAF  & \textbf{0.454} & \textbf{13.41} & \textbf{0.344} & \textbf{16.19} & 14.8\\ \bottomrule[1.5pt]
    \end{tabular}
    \label{tab:ablation2}
\end{table*}

The visualization in Fig. \ref{fig:qualitative} further demonstrates the robustness of the TAF method. In case (a), the first object on the left has a high motion speed relative to the camera, so when using Event Volume taking $\Delta \tau=200ms$ and Surface of Active Events taking $\lambda=1\times10^{-6}$, both cannot detect the object due to the motion blur. Although the object can be detected when using Event Count Image under both $N=50,000$ and $N=200,000$, the estimation of the height is inaccurate. In case (d), the two objects on the right side have high motion speed relative to the camera. Therefore, also due to the motion blur, the estimation of the size is inaccurate when using Event Volume taking $\Delta \tau=200ms$, the localization is inaccurate when using Event Count Image taking $N=200,000$, and the object is not detectable when using Surface of Active Events taking $\lambda=1\times10^{-6}$. On the other hand, in both cases (b)(c)(d), there are objects with low motion speed relative to the camera. It can be seen that when using Event Volume taking $\Delta \tau=50ms$ and Surface of Active Events taking $\lambda=1\times10^{-5}$, the first object on the left in case (b) and the second object on the left in case (c) are not detected, while the size estimation of the first object on the left in case (d) is not accurate. When using Event Count Image taking $N=50,000$, the second object from the left in case (c) is not detected and the size of the first object from the left in case (d) is not estimated accurately. In contrast, the TAF method can detect all the objects mentioned above while estimating the size and location accurately.

\subsection{Ablation Study}

\begin{table*}[htbp]
    \caption{The performance of the TAF under different settings on the GEN1 Dataset. The bold is the best result in each group of comparisons. }
    \centering
    \begin{tabular}{cc|cccccccc}
    \toprule[1.5pt]
        $K$ & BFM & \thead{Lv1\\mAP} & \thead{Lv2\\mAP} & \thead{Lv3\\mAP} & \thead{Lv4\\mAP} & \thead{Lv5\\mAP} & \thead{Overall\\mAP} & \thead{Representation\\Time(ms)} & \thead{Inference\\Time(ms)} \\ \hline
        4 & - & 0.272 & 0.438 & 0.456 & 0.444 & 0.396& 0.444 & \textbf{1.43} & \textbf{11.46}  \\  
        4 & \checkmark & 0.282 & \textbf{0.461} & 0.476 & \textbf{0.459} & \textbf{0.407}& \textbf{0.454} & \textbf{1.43} & 11.98 \\
        8 & - & 0.269 & 0.439 & 0.456 & 0.449 & 0.398& 0.445 & 1.76 & 11.67  \\
        8 & \checkmark & \textbf{0.288} & 0.451 & \textbf{0.478} & 0.457 & 0.402 & 0.451 & 1.76 & 12.26\\ \bottomrule[1.5pt]
    \end{tabular}
    \label{tab:ablation gen1}
\end{table*}

\begin{table*}[htbp]
    \caption{The performance of the TAF under different settings on 1 MEGAPIXEL Dataset (Subset). The bold is the best result in each group of comparisons. }
    \centering
    \begin{tabular}{cc|cccccccc}
    \toprule[1.5pt]
        $K$ & BFM & \thead{Lv1\\mAP} & \thead{Lv2\\mAP} & \thead{Lv3\\mAP} & \thead{Lv4\\mAP} & \thead{Lv5\\mAP} & \thead{Overall\\mAP} & \thead{Representation\\Time(ms)} & \thead{Inference\\Time(ms)} \\ \hline
        4 & - & 0.230 & 0.284 & 0.313 & 0.293 & 0.301 & 0.326 & \textbf{1.76} & \textbf{11.67} \\  
        4 & \checkmark & 0.222 & 0.270 & 0.313 & 0.291 & 0.300 & 0.333 & \textbf{1.76} & 12.84 \\
        8 & - & 0.258 & 0.296 & 0.315 & 0.294 & 0.300 & 0.323 & 2.83 & 12.38 \\
        8 & \checkmark & \textbf{0.262} & \textbf{0.312} & \textbf{0.339} & \textbf{0.308} & \textbf{0.314} & \textbf{0.344} & 2.83 & 13.36 \\ \bottomrule[1.5pt]
    \end{tabular}
    \label{tab:ablation gen4}
\end{table*}

In the ablation experiments, we focus on the effectiveness of the different components of our method and factors that affect the performance when using the TAF representation method. 

\subsubsection{Components}

The ablation of components is shown in TABLE \ref{tab:ablation}, where the runtime is the sum of the representation time and the model inference time. Experiments show that our proposed data augmentation scheme can improve the accuracy of the baseline detector by a large margin on both datasets. With a slight increase in the number of parameters compared to YOLOX, AED notably improves accuracy and significantly reduces runtime on both datasets. Moreover, using the TAF method instead of the Event Volume will result in a significant improvement in accuracy and a reduction of the running time on both datasets. The reduction of the running time is especially significant on the 1 MEGAPIXEL Dataset (Subset). Our final method achieves 74.6 FPS on the Prophesee GEN1 Automotive Detection Dataset (GEN1 Dataset)~\cite{de2020large} and 61.8 FPS on the Prophesee 1 MEGAPIXEL Automotive Detection Dataset ~\cite{perot2020learning}, which satisfies the real-time processing requirements.

We also evaluate the effectiveness of inserting TAF into various SOTA event data object detection methods. As shown in TABLE \ref{tab:ablation2}, our experimental results show that TAF significantly improves the detection accuracy of these methods while maintaining a high inference speed. The results show the generalizability of our proposed TAF method.

\subsubsection{Temporal Active Focus}

TABLE \ref{tab:ablation gen1} and \ref{tab:ablation gen4} show the performance of the TAF under different settings. Overall, using BFM for feature pre-extracting will result in remarkable accuracy improvement at all five motion levels on both datasets. On the 1 MEGAPIXEL Dataset (Subset), we can see a higher performance improvement for detecting objects that are slow relative to the camera. However, the running speed is slightly reduced because feature extraction needs to be performed point-wisely in space. On GEN1 Dataset, the queue depth $K$ value has little impact on accuracy, and we can see that $K=4$ is a better hyper-parameter on GEN1 Dataset. On the 1 MEGAPIXEL Dataset (Subset), when $K=8$, the accuracy is slightly lower without BFM. However, with BFM, the accuracy is much higher. This further illustrates the effectiveness of the BFM module: the 1 MEGAPIXEL Dataset (Subset) has a higher resolution, and the camera used is more sensitive to changes in illumination, requiring larger $K$ values to aggregate events over a larger time range to fully retain event information. However, larger $K$ values bring richer semantics to the temporal dimension. The traditional convolutional input layer is difficult to extract the information. On the other hand, BFM can extract the rich semantics as much as possible, thus fully exploiting the effect of the TAF method. 

\section{Conclusions}

In this paper, we present a motion robust and high-speed detection pipeline for event-based object detection, which takes the different velocities of objects into account and further reduces the computational burden compared with previous event-based object detectors. Specifically, we introduce the  Temporal Active Focus (TAF) event representation, the Bifurcated Folding Module (BFM), the Agile Event Detector (AED), and a simple yet effective data augmentation strategy. The TAF leverages the spatial-temporal asynchronous event data and builds event tensors robust to object motions, the BFM extracts rich temporal information at the input layer, and the AED is faster and more accurate than the baseline YOLOX. Extensive experiments on two typical event-based object detection datasets show that our detection pipeline achieves leading accuracy compared with the state-of-the-art event-based object detector. In addition, our method has a far lower number of parameters and much higher running speed while achieving competitive accuracy and high motion robustness.    

\section*{Acknowledgement}

We would like to thank Mr. Etienne Perot, Senior Machine Learning Engineer from Prophesee, for providing the implementation details of the methods in \cite{perot2020learning}. We would like to thank Mr. Jianing Li, Ph.D. Candidate from Multimedia Learning Group, National Engineering Laboratory for Video Technology, Peking University for providing details about the experimental conditions and the size of the neural network models in \cite{li2022asynchronous}. 

\bibliographystyle{IEEEtran}
\bibliography{mybibfile}

\begin{IEEEbiography}[{\includegraphics[width=1in,height=1.25in,clip,keepaspectratio]{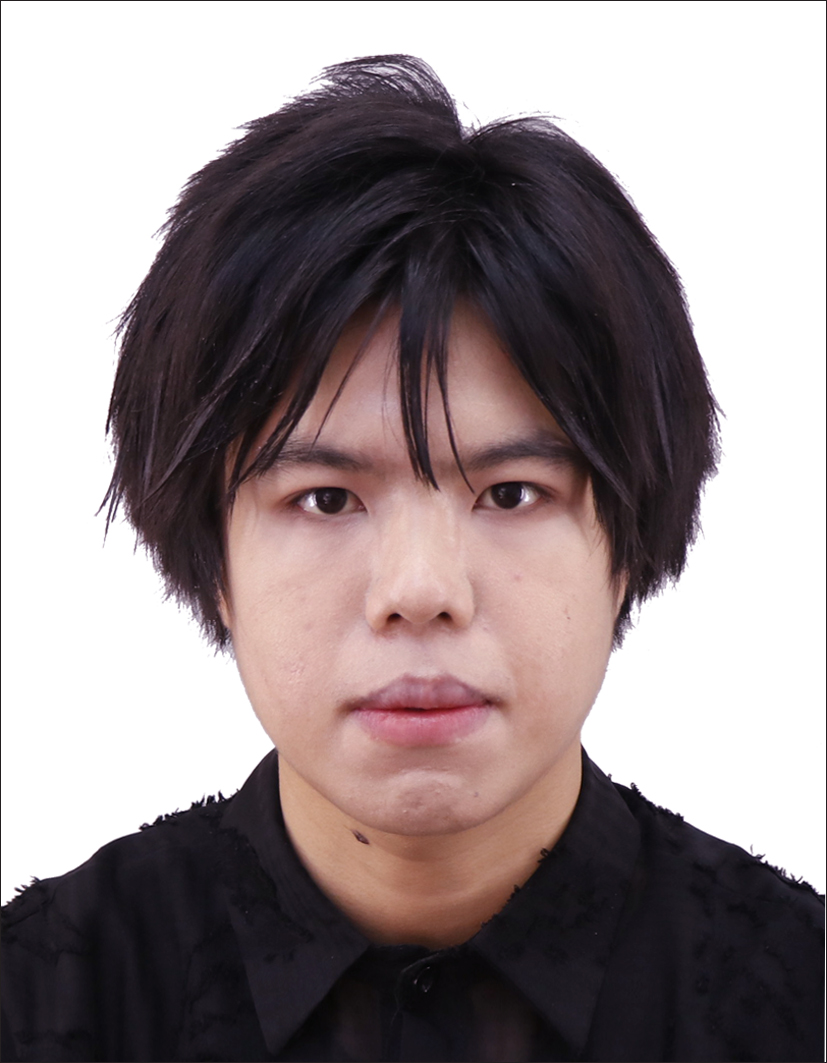}}]{Bingde Liu} received his B.S. degree in Communication Engineering from Wuhan University, Wuhan, China, in 2022, his MSc. degree in Financial Technology with Data Science from University of Bristol, Bristol, U.K., in 2023, he is currently pursuing his Ph.D. degree in Industrial Engineering and Economics at Tokyo Institution of Technology, Tokyo, Japan. His research focuses on the application of artificial intelligence. 
\end{IEEEbiography}

\begin{IEEEbiography}[{\includegraphics[width=1in,height=1.25in,clip,keepaspectratio]{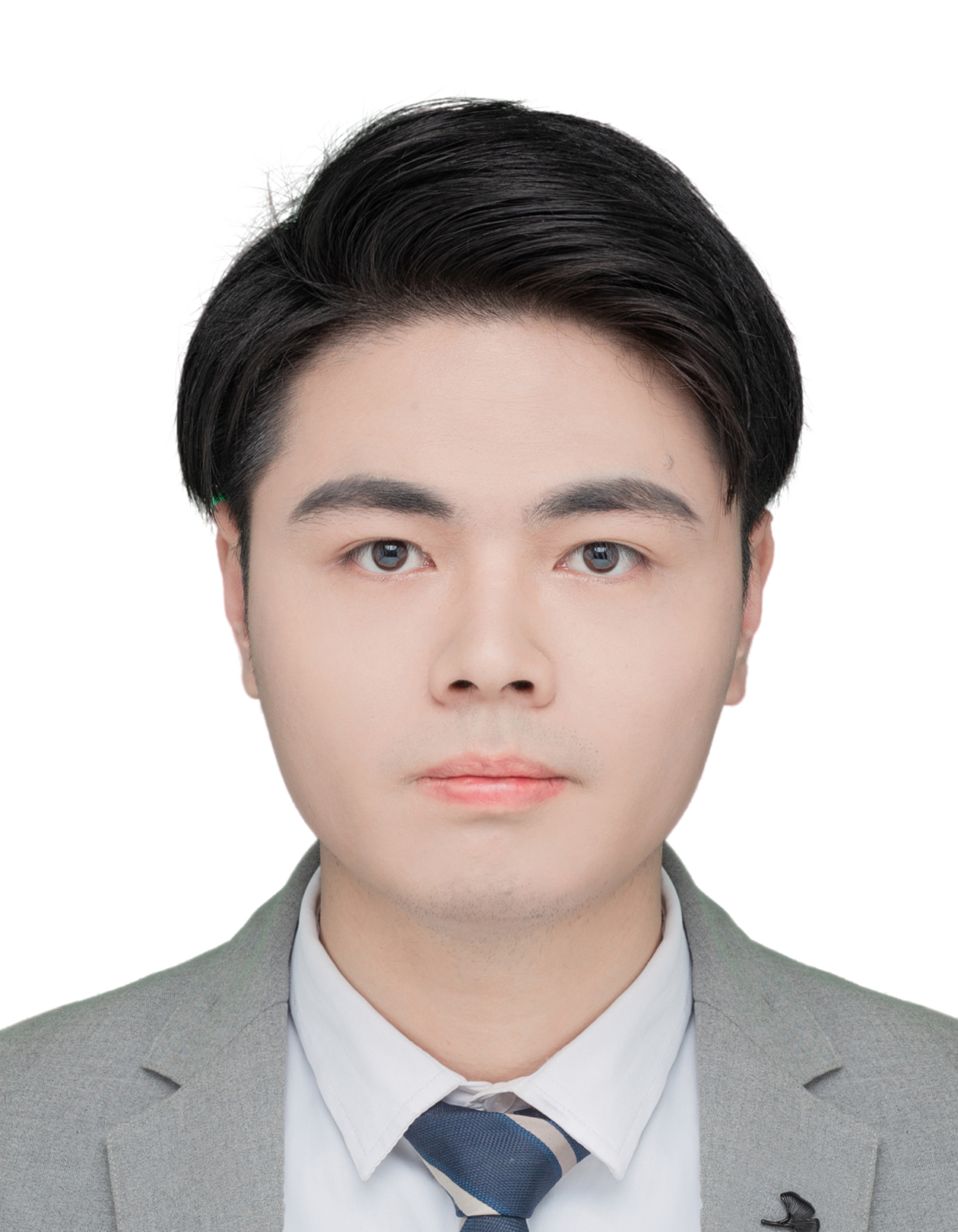}}]{Chang Xu}
received his B.S. degree in electronic information engineering from Wuhan University, Wuhan, China, in 2021, he is currently pursuing his M.S. degree in communication and information system at Wuhan University, Wuhan, China. His research focuses on tiny object detection, oriented object detection, and label noise.
\end{IEEEbiography}

\begin{IEEEbiography}[{\includegraphics[width=1in,height=1.25in,clip,keepaspectratio]{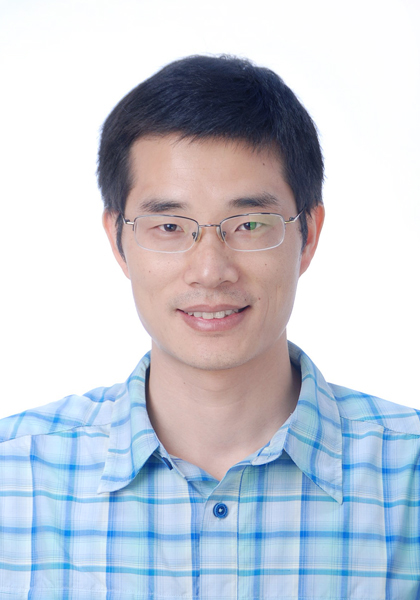}}]{Wen Yang}(Senior Member, IEEE)~received his B.S. degree in Electronic Apparatus and Surveying Technology, his M.S. degree in Computer Application Technology, and his Ph.D. degree in Communication and Information System, all from Wuhan University, Wuhan, China, in 1998, 2001, and 2004, respectively. In 2008 and 2009, he worked as a Visiting Scholar with the Apprentissage et Interfaces (AI) Team at the Laboratoire Jean Kuntzmann in Grenoble, France. Following that, he served as a Post-Doctoral Researcher with the State Key Laboratory of Information Engineering, Surveying, Mapping, and Remote Sensing, also at Wuhan University, from 2010 to 2013. Since then, he has held the position of Full Professor at the School of Electronic Information, Wuhan University. His research interests include object detection and recognition, multisensor information fusion, and remote sensing image interpretation.
\end{IEEEbiography}

\begin{IEEEbiography}
[{\includegraphics[width=1in,height=1.25in,clip,keepaspectratio]{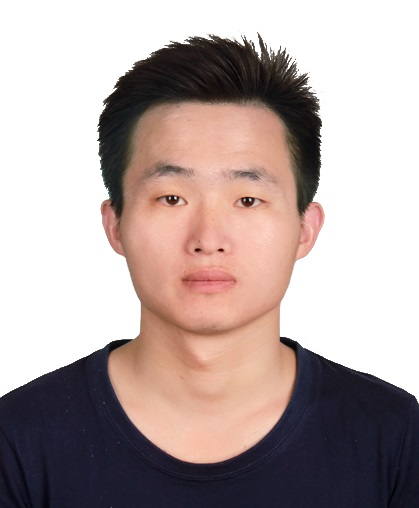}}]{Huai Yu}
(Member, IEEE)~received his B.S. and Ph.D. degrees in communication and information systems from Wuhan University, China, in 2015 and 2020, respectively. From 2018 to 2020 and 2020 to 2021, he has been a visiting scholar and postdoctoral fellow at the Robotics Institute, Carnegie Mellon University, Pittsburgh, PA, USA. He is currently working as a research associate professor at the School of Electronic Information, Wuhan University. His research involves multi-modal visual feature detection and matching, structure from motion, and SLAM.
 \end{IEEEbiography}

\begin{IEEEbiography}[{\includegraphics[width=1in,height=1.25in,clip, keepaspectratio, trim={0 50 0 90}]{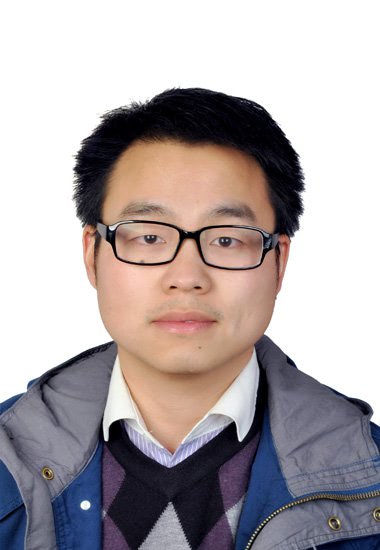}}]{Lei Yu}
received his B.S. and Ph.D. degrees in signal processing from Wuhan University, Wuhan, China, in 2006 and 2012, respectively. From 2013 to 2014, he has been a Postdoc Researcher with the VisAGeS Group at the Institut National de Recherche en Informatique et en Automatique (INRIA) for one and a half years. He is currently working as an associate professor at the School of Electronics and Information, Wuhan University, Wuhan, China. From 2016 to 2017, he was also a Visiting Professor at Duke University for one year. He has been working as a guest professor in the École Nationale Supérieure de l'Électronique et de ses Applications (ENSEA), Cergy, France, for one month in 2018. His research interests include neuromorphic vision and computation.
\end{IEEEbiography}

\vspace{12pt}

\end{document}